\pdfoutput=1
\documentclass[11pt]{article}

\usepackage[final]{acl}

\usepackage{lineno}

\usepackage{times}
\usepackage{latexsym}

\usepackage[T1]{fontenc}

\usepackage[utf8]{inputenc}

\usepackage{microtype}

\usepackage{inconsolata}

\usepackage{graphicx}

\usepackage{xcolor}
\definecolor{customborder}{HTML}{6992f9}
\definecolor{spring_red}{HTML}{FD7F6F} 
\definecolor{spring_blue}{HTML}{7EB0D5} 
\definecolor{spring_green}{HTML}{B2E061} 

\usepackage{tcolorbox}
\tcbuselibrary{listings, skins}

\tcbset{
  colback=gray!5!white,
  colframe=gray!80!black,
  boxrule=0.5pt,
  arc=3pt,
  auto outer arc,
  boxsep=5pt,
  left=4pt,
  right=4pt,
  top=4pt,
  bottom=4pt
}

\usepackage{newfloat}

\DeclareFloatingEnvironment{boxes} 
\newcommand{\boxref}[1]{\hyperref[{#1}]{Box~\ref*{#1}}}

\newcommand{\model}{\textbf{VERBA}}
\usepackage{enumitem} 
\usepackage{color}
\usepackage{soul}
\usepackage{booktabs}
\usepackage{amsmath,amssymb}
\usepackage{subcaption}
\usepackage{multirow}
\usepackage{hyperref}
\usepackage{caption} 
\usepackage{array}
\usepackage{titlesec}
\titlespacing*{\paragraph}{0pt}{\baselineskip}{1em}

\newcommand{\Accmismatch}{\mathbf{Acc}_{\text{mismatch}}}
\newcommand{\Accmatch}{\mathbf{Acc}_{\text{match}}}
\newcommand{\Accoverall}{\mathbf{Acc}_{\text{overall}}}
\newcommand{\accuracyWithError}[2]{%
    \text{\( #1 \)} \scalebox{0.8}{\(\pm #2\)}%
}

\usepackage{amsmath,amsfonts,bm}

\def\eqref#1{equation~\ref{#1}}

\def\1{\bm{1}}

\DeclareMathAlphabet{\mathsfit}{\encodingdefault}{\sfdefault}{m}{sl}
\SetMathAlphabet{\mathsfit}{bold}{\encodingdefault}{\sfdefault}{bx}{n}

\usepackage{hyperref}
\usepackage{url}

\usepackage{listings}
\usepackage{xcolor}

\title{\model{}: Verbalizing Model Differences Using Large Language Models }

\author{Shravan Doda, Shashidhar Reddy Javaji, Zining Zhu \\
  Stevens Institute of Technology \\
  Hoboken, NJ, USA \\}

\begin{document}

\maketitle

\begin{abstract}
In the current machine learning landscape, we face a ``model lake'' phenomenon: Given a task, there is a proliferation of trained models with similar performances despite different behavior. For model users attempting to navigate and select from the models, documentation comparing model pairs is helpful. However, for every $N$ models there could be $\mathcal{O}(N^2)$ pairwise comparisons, a number prohibitive for the model developers to manually perform pairwise comparisons and prepare documentations. 
To facilitate fine-grained pairwise comparisons among models, we introduced \model{}. Our approach leverages a large language model (LLM) to generate verbalizations of model differences by sampling from the two models.
We established a protocol that evaluates the informativeness of the verbalizations via simulation. We also assembled a suite with a diverse set of commonly used machine learning models as a benchmark. 
For a pair of decision tree models with up to 5\% performance difference but 20-25\% behavioral differences, \model{} effectively verbalizes their variations with up to 80\% overall accuracy. When we included the models' structural information, the verbalization's accuracy further improved to 90\%.
\model{} opens up new research avenues for improving the transparency and comparability of machine learning models in a post-hoc manner.
\end{abstract}

\section{Introduction}

The rapid increase in the number of machine learning models across various domains has led to a ``model lake'' phenomenon \citep{pal2024modellakes}: navigating and selecting models has been increasingly challenging. It's often a struggle to discern the strengths and weaknesses and identify the most appropriate model for a task. 

Several efforts have been made to improve model management and documentation. One example is ModelDB \citep{vartak2016modeldb}, which serves as a versioning system that tracks models' metadata across successive iterations (such as model configurations, training datasets, and evaluation metrics). ModelDB’s primary focus is on ensuring reproducibility and traceability of models over time, allowing users to track changes and reproduce past experiments. Similarly, Model Cards \citep{Mitchell_2019} and Data Cards \citep{pushkarna2022datacardspurposefultransparent}, along with recent work on their automated generation \citep{liu2024automaticgenerationmodeldata}, offer valuable documentation on data characteristics, model architectures, and training processes. 

However, model selection is challenging even if each model is well-documented and well-evaluated with corresponding performance metrics, because model selection involves pairwise comparison. Frequently, two models have almost the same performance metrics, but their behaviors differ drastically. While the methods documenting each model individually provide critical insights into individual models and datasets, they do not directly explain the fine-grained differences between the models' behavior. Research aimed at systematically differentiating models remains sparse, calling for a more transparent model comparison framework.

Recently, large language models (LLMs) have shown exceptional capabilities over a diverse range of tasks \citep{hendy2023good,brown2020language}. Among these capabilities, the potential of LLMs to explain model behavior is especially intriguing \citep{kroeger_are_2023,singh2023Explaining}. Motivated by their works, we built a framework, \model{}, that leverages LLMs to verbalize the model differences. 

The \model{} framework is designed to compare two models trained on the same dataset by verbalizing their differences. It does so by serializing a representative sample of input instances (from the dataset) and the corresponding model outputs in a JSON format. The serialization, along with a task description, is passed to the LLM through a zero-shot-based prompt. The LLM then analyzes the patterns from the serialization, captures the inconsistencies in the predictions between the two models, and summarizes them in natural language. 
The \model{} framework is flexible. Since the framework primarily relies on comparing input-output samples, it can be used with various model types and datasets. Additionally, \model{} is extensible. The framework allows the user to incorporate model-specific information, for example, textual descriptions of the structures of decision trees, which can improve the informativeness of the verbalization --- we present the effects via ablation studies in \autoref{subsec:ablation-model-internals} and \autoref{subsec:ablation-model-type}. 

To systematically evaluate the verbalizations of \model{}, we established a protocol inspired by the evaluation of natural language explanations \citep{kopf_cosy_2024,singh2023Explaining}. Given the inputs, the first model's outputs, and the verbalization, we use an external LLM to infer the second model's outputs. The accuracy of the inference (averaged after exchanging the two models) is a proxy of the informativeness of the verbalization.

To complement the evaluation protocol, we created a benchmark containing three types of commonly used AI models: logistic regressors, decision trees, and multilayer perceptrons. The models were systematically perturbed so that, despite resembling each other in performance, their behavior differed. We stratified the models based on their behavioral differences. 

On these models, we benchmarked \model{} utilizing state-of-the-art LLMs across three machine learning datasets. \model{} was able to achieve over 80\% match accuracy over model pairs that have $\leq 5\%$ accuracy differences. 

Our work provides a valuable starting point for using LLMs to verbalize the behavior differences of AI models, enhancing their transparency and comparability in a post-hoc manner. Our work can be a critical building block for the automated selection and management of AI models.

\section{Related Works}

\paragraph{Neuron-Level Explanations}
Research into the semantics of individual DNN components, particularly neurons, has evolved significantly. Early investigations, such as those by \citet{mu_compositional_2022}, focused on identifying compositional logical concepts within neurons. Building on this, \citet{hernandez_natural_2022} developed techniques to map textual descriptions to neurons by optimizing pointwise mutual information. More recent approaches have incorporated external models to enhance explanations of neuron functions. For instance, \citet{bills_language_2023} conducted a proof-of-concept study using an external LLM, such as GPT-4, to articulate neuron functionalities. However, the perfection of these methods remains elusive, as noted by \citet{huang_rigorously_2023}. Evaluating the effectiveness of these explanations is currently a vibrant area of inquiry, with ongoing studies like those by \citet{kopf_cosy_2024} and \citet{mondal_towards_2024}.


\paragraph{Model-Level Explanations}
Beyond individual neurons, the field is extending towards automated explanation methods for broader model components. \citet{singh2023Explaining} approaches models as opaque ``text modules'', providing explanations without internal visibility. Our methodology diverges by incorporating more detailed information about the models, which we believe enhances the accuracy of explanations, an approach recommended by \citet{ajwani2024LLMgenerated}. Notably, our work aligns with \citet{kroeger_are_2023}, who employ in-context learning for prompting LLMs to explain machine learning models. Our strategy differs as we emphasize zero-shot instructions.

\paragraph{Behavioral Comparison Framework}
A related approach in behavioral comparison is \textsc{BehaviorBox} by \citet{tjuatja2025behaviorboxautomateddiscoveryfinegrained}, a framework that automatically discovers fine-grained features where the performance of the two language models diverges. Using a sparse autoencoder over performance-aware embeddings, it extracts and labels coherent data slices, such as specific syntactic patterns of formatting, that cause performance gaps. While \textsc{BehaviorBox} focuses on the structured discovery of what to explain, our \model{} framework generates natural language verbalizations of how the models' behaviors differ based on their outputs.

\paragraph{Extract Interpretable Features}
Concurrently, an avenue of research in mechanistic interpretability extracts interpretable features directly from neurons. Techniques such as learning sparse auto-encoders have been explored by \citet{bricken2023monosemanticity}. A significant advancement by \citet{templeton_scaling_2024} scales up these efforts to newer architectures like Claude 3.5 Sonnet \citep{introducingClaude3.5}. Unlike previous methods, we do not assume a predefined set of features for explanation, opting instead to use the LLM as a dynamic verbalizer to generate explanatory content.

Another prevalent mechanistic interpretability approach is the use of the language model head of DNNs as a ``logit lens``, as demonstrated by \citet{nostalgebraist_interpreting_2020}. This method has been further developed and diversified by researchers like \citet{pal_future_2023} and \citet{belrose_eliciting_2023}. The PatchScope framework by \citet{ghandeharioun_patchscopes_2024} extends these techniques, incorporating methods that modify the representations themselves. In our research, rather than utilizing the language model head directly, we employ an external LLM to serve as the verbalizer, providing a novel means of interpreting and explaining model behaviors. 

\paragraph{LLM Distinction}
Several approaches have emerged to differentiate between LLMs. One method, LLM Fingerprinting, introduces a cryptographically inspired technique called Chain and Hash \citep{russinovich2024heythatsmodelintroducing}. This approach generates a set of unique questions (the ``fingerprints'') and corresponding answers, which are hashed to prevent false claims of ownership over models. Complementing this, another method \citep{richardeau202420questionsgamedistinguish} proposes using a sequence of binary questions, inspired by the 20 Questions game, to determine if two LLMs are identical.
Unlike fingerprinting or binary distinction, our framework focuses on the behavioral aspect of models. Moreover, our current work does not aim to compare LLMs themselves; rather, we leverage LLMs as a tool to compare and verbalize the differences among other models. 
Similarly, \citet{zhong2022describing} described the difference between text distributions. We describe the differences between AI models, including but not limited to those based on text.

\paragraph{Model Selection and Ranking}
A recent approach \citep{okanovic2024modelswrongusefulmodel} introduces ``Model Selector'', a framework for label-efficient selection of the best-performing model among pretrained classifiers. 
Similarly, \citet{you2022model_hub} ranked the models within a model hub. \citet{ong2024routellm} developed a routing mechanism that selects the appropriate model, and \citet{frick2025prompt} showed that the approaches based on model hubs could lead to systems with superior performance. Our work is also relevant to improving the models in a model hub, but we focus on assessing and interpreting the model behavior.

\paragraph{LLM-driven Hypothesis Generation} Recent work by \citet{Zhou_2024} proposes HypoGeniC, a framework that uses LLMs to generate interpretable scientific hypotheses from small labeled datasets. Our \model{} framework can also be considered a type of hypothesis generation framework, where our data are sampled from the machine learning models, and our hypotheses are the verbalization of model differences. 

\begin{figure*}[t] 
  \centering
  \includegraphics[width=0.85\linewidth]{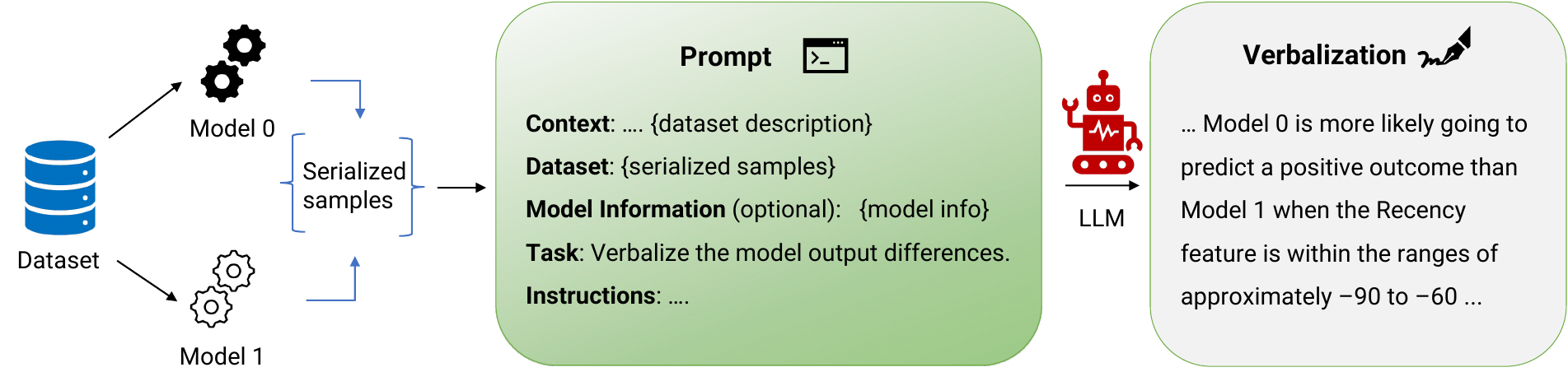}
  \caption{Overview of the \model{} framework. Given a dataset and a pair of models trained on a dataset, \model{} verbalizes the differences between the two models.}
  \label{fig:framework-overview}
\end{figure*}

\section{The \model{} Framework}
\label{sec:repr_sample}

\vspace{-10pt}

Here we present our \model{} framework. As illustrated by \autoref{fig:framework-overview}, \model{} generates the verbalizations, i.e., natural-language descriptions of the differences between two machine learning models trained on the same dataset. 

\paragraph{Notation:} Let $\mathbf{X} = \{\mathbf{x_i}\}_{i=1}^n$ be a tabular dataset where each $\mathbf{x_i} \in \mathbb{R}^d$ represents a feature vector. Since we considered classification tasks, suppose the target vector is $\mathbf{y} = \{y_i\}_{i=1}^n$, where $y_i \in C$ and $C$ is a set of possible classes. We denote a subset of the dataset as $\mathbf{X}_{\text{sub}}$, with size $n_{\text{sub}}$. Similarly, the corresponding subset of target values is denoted by $\mathbf{y}_{\text{sub}} = \{y_i\}_{i=1}^{n_{\text{sub}}}$. We define the feature names of $\mathbf{X}$ as $F = \{f_1, f_2, \dots, f_d\}$, where each $f_i$ represents a natural-language description of a feature, such as ``age'' or ``glucose''.

Let $M_0$ and $M_1$ be the two models that we compare with \model{}. For each data point $\mathbf{x_i} \in \mathbf{X}_{\text{sub}}$, the predicted target values from models $M_0$ and $M_1$ are represented as $\hat{y}^{(0)}_{\text{sub},i} = M_0(\mathbf{x_i})$ and $\hat{y}^{(1)}_{\text{sub},i} = M_1(\mathbf{x_i})$, respectively. The corresponding predicted target vectors for the subset are denoted by $\hat{\mathbf{y}}^{(0)}_{\text{sub}}$ and $\hat{\mathbf{y}}^{(1)}_{\text{sub}}$.

\paragraph{Representative Sample:} We constructed our representative sample using the $verb$ split of the dataset $\mathbf{X_\text{verb}}$ (size $n_{\text{verb}}$) along with the predicted target vectors $\mathbf{\hat{y}}^{(0)}_{\text{verb}}$ and $\mathbf{\hat{y}}^{(1)}_{\text{verb}}$ from models $M_0$ and $M_1$ respectively. Before passing the verbalization sample \{${\mathbf{X_\text{verb}}, \mathbf{\hat{y}^{(0)}_{\text{verb}}}, \mathbf{\hat{y}^{(1)}_{\text{verb}}}}$\} to the LLM, we serialize it into a JSON format.

\paragraph{LLM for Verbalization:} The framework can be used with different LLMs. Let $LLM_{\text{verb}}$ represent the LLM responsible for generating verbalizations. The verbalization produced, denoted by $\mathbf{v}$, lies within the vocabulary space of $LLM_{\text{verb}}$.

\paragraph{Prompt:} We assembled the serialized results into a prompt to the verbalizer $LLM_{\text{verb}}$. Our prompt was inspired by previous LLM work in explainable AI \citep{kroeger_are_2023} and included the following elements: \textbf{\textit{Context}}, \textbf{\textit{Dataset}}, \textbf{\textit{Model Information}}, \textbf{\textit{Task}}, and \textbf{\textit{Instructions}}. 

The \textbf{\textit{Context}} outlines the type of models used, their classification task, and a general overview of the dataset, including details about the features and the target variable. We chose to explicitly mention the feature names, $F = \{f_1, f_2, \dots, f_d\}$, drawing insights from previous work \citep{hegselmann2023tabllmfewshotclassificationtabular}, which showed that feature names can help improve interpretability. We included the order of features in the representative sample to ensure that $LLM_{\text{verb}}$ can correctly associate feature names with their corresponding feature values. Additionally, we explicitly explained the meaning of the target variable, including what each possible value $c \in C$ represents. 

The \textbf{\textit{Dataset}} was the serialized representative sample, as described above.

The \textbf{\textit{Model Information}} was kept optional. It could describe either the structural information or just the names of the two models. We studied the effects of the two choices in ablation studies (\autoref{sec:experiment-results}.

The \textbf{\textit{Task}} section stated the underlying task we wanted $LLM_{\text{verb}}$ to perform, i.e., generate verbalization of the decision boundaries of the two models based on the samples. 

The \textbf{\textit{Instructions}} enumerated detailed instructions for the LLM. These included: analyzing where the outputs of the two models diverged and aligned, and describing the specific ranges of feature values in the decision boundaries. We also instructed the verbalizer to identify the key contributing features.

An assembled prompt is illustrated in \autoref{prompts}.

\section{Evaluation} If a verbalization $\mathbf{v}$ accurately captured the differences between two models, it was expected to facilitate an evaluator to predict the second model's outputs given the inputs and the outputs of the first.

We used an LLM to be the evaluator, and referred to it as $LLM_{\text{eval}}$. It received the verbalization $\mathbf{v}$ along with an evaluation sample $\{\mathbf{X_{eval}}, \hat{y}^{(0)}_{eval}\}$ to simulate $\tilde{y}^{(1)}_{eval}$, and vice versa. If we let $k \in \{0,1\}$ to index the two models, we have:
\begin{equation}
\tilde{y}^{(k)}_{\text{eval}} = LLM_{\text{eval}} (\mathbf{X_{\text{eval}}}, \hat{y}^{(1-k)}_{\text{eval}}\,;\,\mathbf{v})
\label{eq:simulation}
\end{equation}
To assess the accuracy of simulated output $\tilde{y}^{(k)}_{\text{eval}}$, we compared it with the original model output $\hat{y}^{k}_{\text{eval}}$ using the following three evaluation metrics: 

\paragraph{Mismatch Accuracy ($\Accmismatch$)} 
It evaluates the instances where the outputs of $M_0$ and $M_1$ disagree for $\mathbf{X_{eval}}$, i.e., $I_{\text{mismatch}} = \left\{ i \mid \hat{y}_{\text{eval}, i}^{(0)} \neq \hat{y}_{\text{eval}, i}^{(1)} \right\}$. For these instances, the metric is computed as the proportion of cases where the simulated output of $M_k$ matches that of the original output of $M_k$, i.e., 
\begin{align*}
&\Accmismatch = \\
&\frac{\sum_{k\in \{0,1\}}\left| \left\{ i \mid \tilde{y}^{(k)}_{\text{eval},i} = \hat{y}^{(k)}_{\text{eval},i}, i\in I_{\text{mismatch}} \right\} \right| }{| I_{\text{mismatch}}|}.
\end{align*}
The $\Accmismatch$ quantifies how well the verbalization $\mathbf{v}$ captures the points of divergence between the two models.

\paragraph{Match Accuracy ($\Accmatch$)} 
It considers the instances where the outputs of $M_0$ and $M_1$ agree, i.e., $I_{\text{match}} = \left\{ i \mid \hat{y}_{\text{eval}, i}^{(0)} = \hat{y}_{\text{eval}, i}^{(1)} \right\}$. The accuracy is similarly computed as the proportion of these cases where the simulated output of $M_k$ matches that of the original output of $M_k$:
\begin{align*}
&\Accmatch = \\
&\frac{\sum_{k\in \{0,1\}}\left| \left\{ i \mid \tilde{y}^{(k)}_{\text{eval},i} = \hat{y}^{(k)}_{\text{eval},i}, i\in I_{\text{match}} \right\} \right| }{| I_{\text{match}}|}.
\end{align*}
The $\Accmatch$ quantifies the extent to which the verbalization $\mathbf{v}$ captures the points of agreement between the two models.

\paragraph{Overall Accuracy (\(\Accoverall\))} It evaluates $\mathbf{v}$'s performance across all instances, combining the cases in both $I_{\text{match}}$ and $I_{\text{mismatch}}$:
\begin{align*}
\Accoverall = \frac{\sum_{k\in \{0,1\}}\left| \left\{ i \mid \tilde{y}^{(k)}_{\text{eval},i} = \hat{y}^{(k)}_{\text{eval},i} \right\} \right| }{| I_{\text{match}}| + |I_{\text{mismatch}}|}.
\end{align*}

To obtain a single score for each metric, we computed it in both directions (simulating $M_1$ from $M_0$ and vice versa) and reported the mean, as written in the summations over $k$ in the above equations. This way, we ensured that the evaluation reflected the verbalization's ability to capture the differences symmetrically between the two models. The evaluation prompt template can be found in \autoref{prompts}.

\section{Experimental Setup}
\label{sec:experiments}

\paragraph{Datasets:} We considered classification tasks, and based on prior work involving LLMs (\citep{hegselmann2023tabllmfewshotclassificationtabular}), we selected the following three datasets: \textbf{Blood} (784 rows, 4 features, 2 classes), \textbf{Diabetes} (768 rows, 8 features, 2 classes), and \textbf{Car} (1,728 rows, 6 features, 4 classes). The datasets were first divided into training and test sets. From the test set, we further split the data into two subsets in a 2:1 ratio: the $verb$ split, which is used as a representative sample for verbalization (as explained in \autoref{sec:repr_sample}), and the $eval$ split, which is reserved for evaluation purposes. This ensured that verbalization and evaluation operate on distinct subsets.

To keep the input context manageable and ensure that each dataset had enough samples in both $verb$ and $eval$ splits, we adjusted the proportions of the initial train-test split. The train-test splits are shown in \autoref{train_test_split_table}.

\begin{table}[h!]
\centering
\resizebox{.4\linewidth}{!}{
\begin{tabular}{lcc}
\toprule
\textbf{Dataset} & \textbf{Train Split (\%)} & \textbf{Test Split (\%)} \\
\midrule
Blood    & 70\%  & 30\%  \\
Diabetes & 70\%  & 30\%  \\
Car      & 87\%  & 13\%  \\
\bottomrule
\end{tabular}}
\caption{Train-test split percentages for datasets}
\label{train_test_split_table}
\end{table}

The datasets were scaled, and preprocessing steps were consistent across all model types.

\paragraph{Models types:} We evaluated our framework on three representative models that span a range of representational capacities: (i) Logistic Regression (LR), (ii) Decision Tree (DT), and (iii) a Multilayer Perceptron (MLP) with a single hidden layer. LR and DT were selected because they are widely used, interpretable, and serve as strong baselines.  To assess the framework’s ability to handle more complex models, we also include a single-layer MLP, which introduces non-linearity. 


\paragraph{Model pairs:} Models have similar performances (in accuracy), but their predictions differ. Such a discrepancy is not reflected by the performance metrics themselves, and we systematically study the utility of \model{} for these models. For each model type, multiple pairs of models ($M_1$ and $M_2$) were trained. The accuracy differences between $M_1$ and $M_2$ are $\le 5\%$. Then, we stratified the experiments based on the percentage of differing outputs between $M_1$ and $M_2$, dividing them into three levels - (i) Level 1 ($15-20\%$), (ii) Level 2 ($20-25\%$), and (iii) Level 3 ($25-30\%$). 


To generate pairs of LR models with a specific percentage of differing outputs, we first trained a base model using \texttt{RandomizedSearchCV} over a broad hyperparameter space. We then created multiple variations by adding randomly generated noise to the base model's coefficients. The noise was controlled by a modification factor $m$ (noise $\sim \mathcal{N}(0, m) $) and applied multiplicatively as $\boldsymbol{\beta} \cdot (1 + \text{noise})$, where $\boldsymbol{\beta}$ denotes the vector of the base model's coefficients. We carefully adjusted $m$ until the percentage of differing outputs between the base and modified models reached the desired level. Rather than limiting our comparisons to the base model obtained from \texttt{RandomizedSearchCV}, we also compared the modified models against each other, identifying a diverse collection of model pairs. We also ensured that the difference in accuracy between any two models in a pair was $ \leq 5\%$, since in practice, models being compared would usually have similar performance.

We follow a similar process for Decision Trees and MLPs, with the details provided in \autoref{add_details:experiments}. For each model type and across all levels of output differences, we generate multiple base models and corresponding modified models.

\paragraph{Verbalizers:} We include four state-of-the-art LLMs as $LLM_{\text{verb}}$: Claude 3.5 Sonnet \citep{introducingClaude3.5}, Gemini 2.0 Flash  \citep{introducingGemini}, GPT-4o \citep{introducingGPT4o} and Llama 3.3 70B \citep{introducingLlama3_3}. For each of these LLMs,  we set the temperature as $T = 0.1$ in their respective API calls.

\paragraph{Evaluator:} For $LLM_\text{eval}$, we use Llama 3.3 70B as a fixed evaluator to avoid evaluator bias and ensure consistent comparison across verbalization models.

\section{Experiment Results}
\label{sec:experiment-results}

\begin{figure}[t]
    \centering
    \begin{subfigure}[t]{\linewidth}
        \centering
        \includegraphics[width=0.9\linewidth]{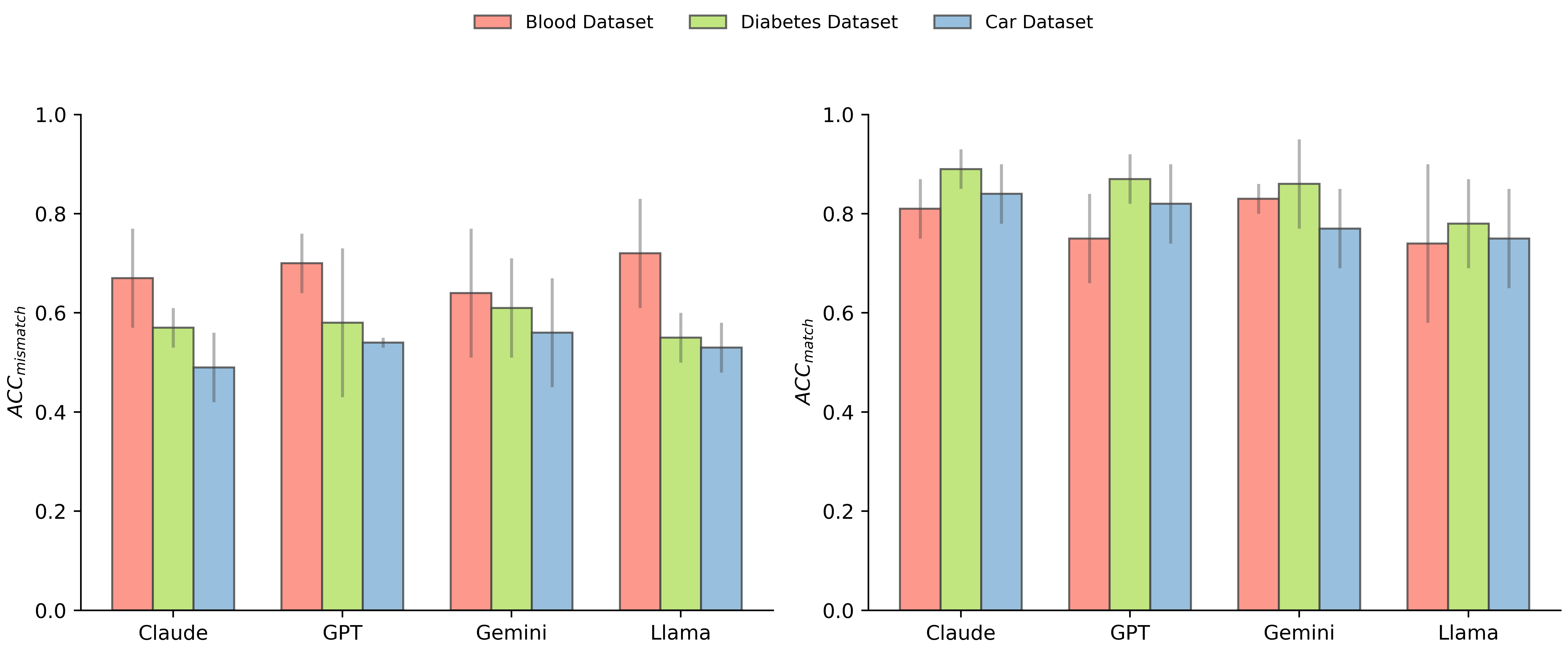}
        \caption{Logistic Regression}
        \label{fig:LRPlot}
    \end{subfigure}
    \hfill
    \begin{subfigure}[t]{\linewidth}
        \centering
        \includegraphics[width=0.9\linewidth]{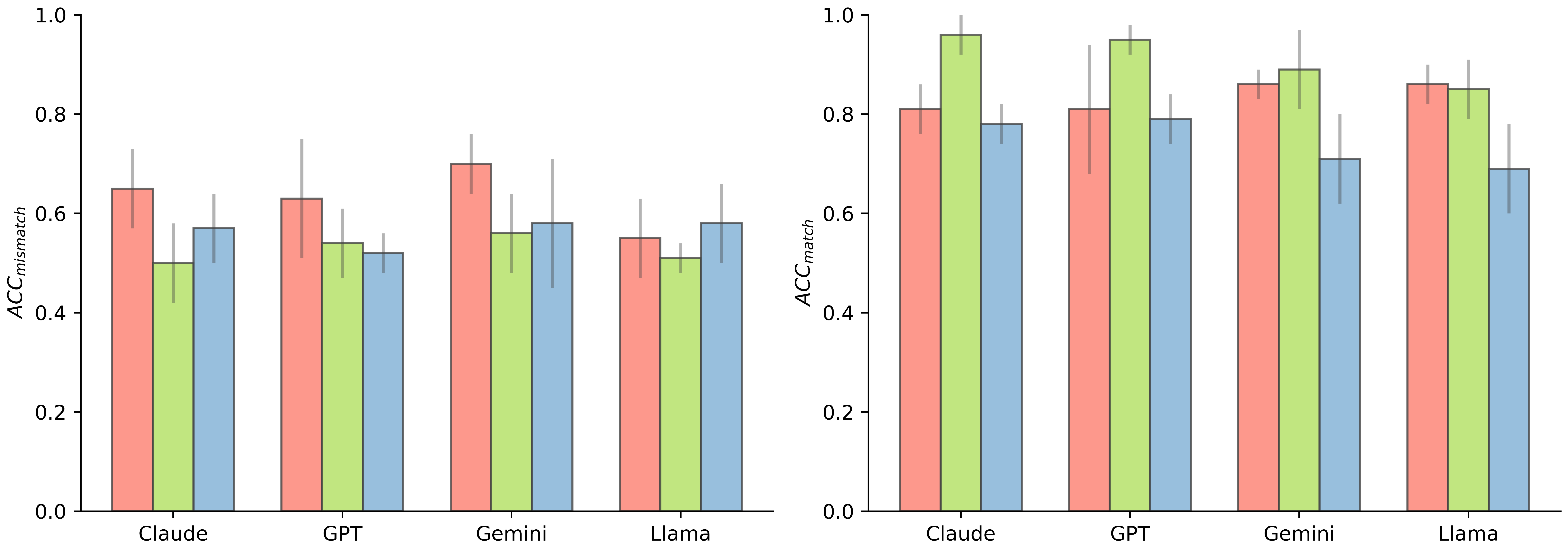}
        \caption{Decision Trees}
        \label{fig:DTPlot}
    \end{subfigure}
    \hfill
    \begin{subfigure}[t]{\linewidth}
        \centering
        \includegraphics[width=0.9\linewidth]{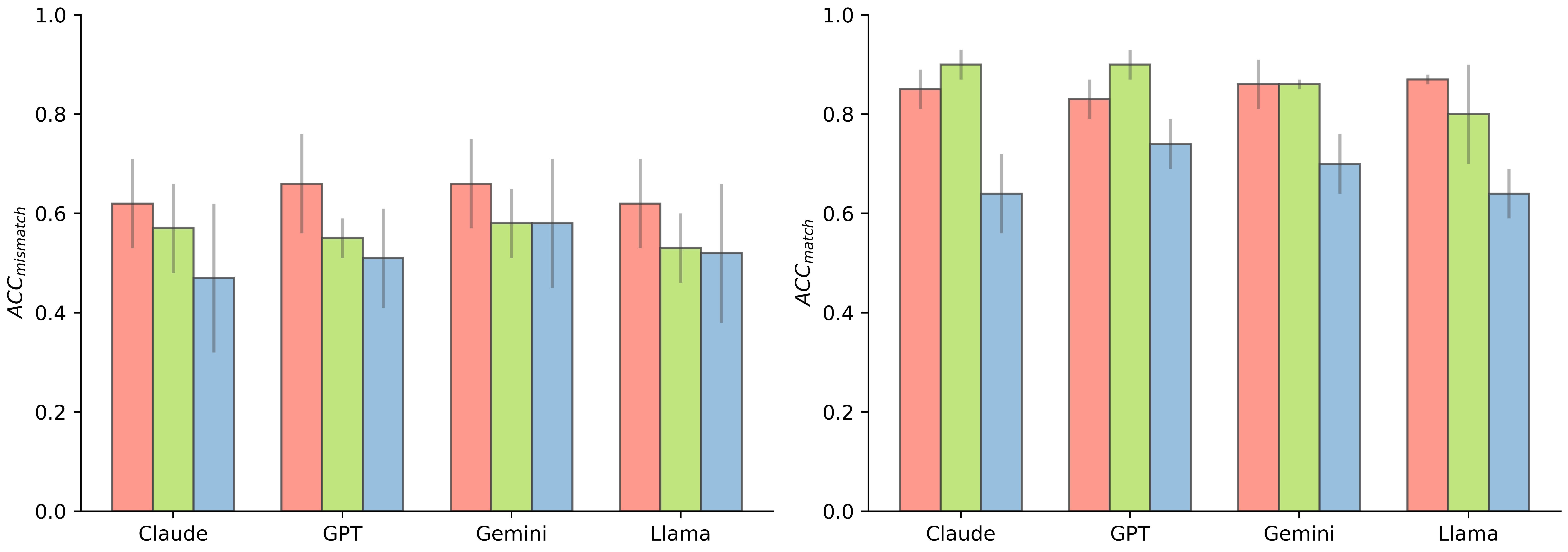}
        \caption{Multilayer Perceptron}
        \label{fig:MLPPlot}
    \end{subfigure}
    \caption{Performance of three LLMs. \ref{fig:LRPlot} shows the \(\Accmismatch\) and \(\Accmatch\) for Level 2 $(20-25\%)$ LR models trained on \textcolor{spring_red}{Blood}, \textcolor{spring_green}{Diabetes}, and \textcolor{spring_blue}{Car} datasets. \ref{fig:DTPlot} and \ref{fig:MLPPlot} shows the same for DTs and MLPs respectively.}
    \label{fig:LRnDTPlot}
\end{figure}


\subsection{Comparing Logistic Regressors}
Our framework demonstrated strong performance when applied to logistic regression (LR) across datasets, likely due to the models' linear nature. \autoref{fig:LRPlot} illustrates the performance on LR models trained on the Blood, Diabetes, and Car datasets. As shown in \autoref{tab:lr_extend}, the $LLM_{\text{verb}}$ performance remained consistent across Levels 1--3 for the Blood dataset.  Gemini outperformed the other LLMs in $\Accoverall$ by a small margin, achieving an $\Accmismatch$ of $\accuracyWithError{0.64}{.13}$ and an $\Accmatch$ of $\accuracyWithError{0.83}{.03}$ on Level 2 $(20-25\%)$ models. GPT-4o and Claude followed closely, with $\Accmismatch$ values of $\accuracyWithError{0.70}{.06}$ and $\accuracyWithError{0.67}{.10}$, and $\Accmatch$ values of $\accuracyWithError{0.75}{.09}$ and $\accuracyWithError{0.81}{.06}$, respectively.

Performance decreased across all datasets at the most challenging level, Level 3 $(25-30\%)$, as detailed in \autoref{tab:lr_extend}. This suggests that as the problem complexity increased, even the best-performing LLMs could not maintain the same level of accuracy.

For the Diabetes and Car datasets, we observed a decline in the performance of the framework, which could be attributed to the increasing complexity of the datasets—Diabetes with a larger number of features and Car with multiple classes. Nevertheless, all three—GPT--4o, Claude, and Gemini—achieved an $\Accoverall$ of $\accuracyWithError{0.75}{.06}$, $\accuracyWithError{0.75}{.05}$, and $\accuracyWithError{0.71}{.05}$ respectively for the Car dataset, which remained substantially above the random-guessing baseline. These results suggest that LLMs were effective at verbalizing differences between logistic regression models. \autoref{tab:verbalization-lr} presents excerpts from some of these verbalizations.


\subsection{Comparing Decision Trees}
Decision Trees posed a greater challenge than LR models, primarily due to their non-linear decision boundaries. Consequently, the framework's performance on DTs was lower, although trends similar to those observed with LR models persisted. Detailed results across all levels and datasets are provided in \autoref{tab:dt_extend}.

\autoref{fig:DTPlot} illustrates that, on the Blood dataset, all four LLMs performed well, with Gemini achieving the highest $\Accmismatch$ of $\accuracyWithError{0.70}{.06}$ and an $\Accmatch$ of $\accuracyWithError{0.83}{.02}$ for Level 2 ($20-25\%$) models. Claude followed closely, with an $\Accmismatch$ of $\accuracyWithError{0.65}{.08}$ and an $\Accmatch$ of $\accuracyWithError{0.77}{.04}$. GPT-4o and Llama also exhibited comparable performance.

The Car dataset introduced additional complexity. Gemini's performance dropped, with an $\Accmismatch$ of $\accuracyWithError{0.58}{.13}$ and an $\Accmatch$ of $\accuracyWithError{0.71}{.09}$. Claude followed closely, achieving an $\Accmismatch$ of $\accuracyWithError{0.57}{.08}$ and an $\Accmatch$ of $\accuracyWithError{0.78}{.04}$.

Despite the decline in overall performance for DTs across datasets, Gemini and Claude managed to maintain relatively strong results. These findings suggest a broader trend: LLMs were generally able to verbalize differences between decision tree models effectively. \autoref{tab:verbalization-dt} presents excerpts from some of these verbalizations.


\subsection{Comparing Multilayer Perceptrons}
Multilayer Perceptrons (MLPs) are more complex than both LR and DTs, and thus posed a greater challenge for our framework. The non-linear nature of MLPs made it more difficult for LLMs to verbalize their differences effectively. Full breakdowns of LLM performance on MLPs can be found in \autoref{tab:mlp_extend}.

\autoref{fig:MLPPlot} illustrates that, on the Blood dataset, all four LLMs performed consistently, with Gemini and Llama emerging as the top performers. Gemini achieved an $\Accmismatch$ of $\accuracyWithError{0.66}{.09}$ and an $\Accmatch$ of $\accuracyWithError{0.86}{0.05}$, while Llama attained an $\Accmismatch$ of $\accuracyWithError{0.62}{.09}$ and an $\Accmatch$ of $\accuracyWithError{0.87}{0.01}$ on the Level 2 $(20-25\%)$ models. GPT-4o and Claude exhibited comparable performance.

On the Car dataset, performance dropped as expected, with Gemini achieving an $\Accmismatch$ of $\accuracyWithError{0.58}{0.13}$ and an $\Accmatch$ of $\accuracyWithError{0.70}{0.06}$. The other LLMs exhibited a similar decline, although their scores remained well above the random-guessing baseline.

Despite the added complexity of MLPs, the framework still achieved reasonable performance. These results suggest that while MLPs are more difficult for LLMs to verbalize, the framework can nonetheless generate meaningful insights into their differences.


\subsection{Ablation on Information about Model's Internals}
\label{subsec:ablation-model-internals}

Access to model internals, compared to solely relying on the representative samples, may help $LLM_{\text{verb}}$ understand (and therefore verbalize) how the models make decisions. We hypothesized that providing such model-specific information would enable LLMs to generate more accurate and faithful verbalizations. We examined the effect of incorporating the models' internals on the performance of our framework in generating verbalizations. By internals, we refer to the textual descriptions of a model's learned structure or information about its inner workings. Different model types expose different internal signals that can inform their predictions. For Logistic Regression (LR), this entailed providing the framework with the learned weights and intercepts. For Decision Trees (DT), we supplied a textual representation of the model structure, focusing on the decision rules and splits. For Multilayer Perceptrons (MLPs), we included the architecture specifications---such as the number of layers, nodes, and activation functions---alongside the learned weights and biases for both the hidden and output layers.

\begin{figure}[t]
    \centering
    \includegraphics[width=0.9\linewidth]{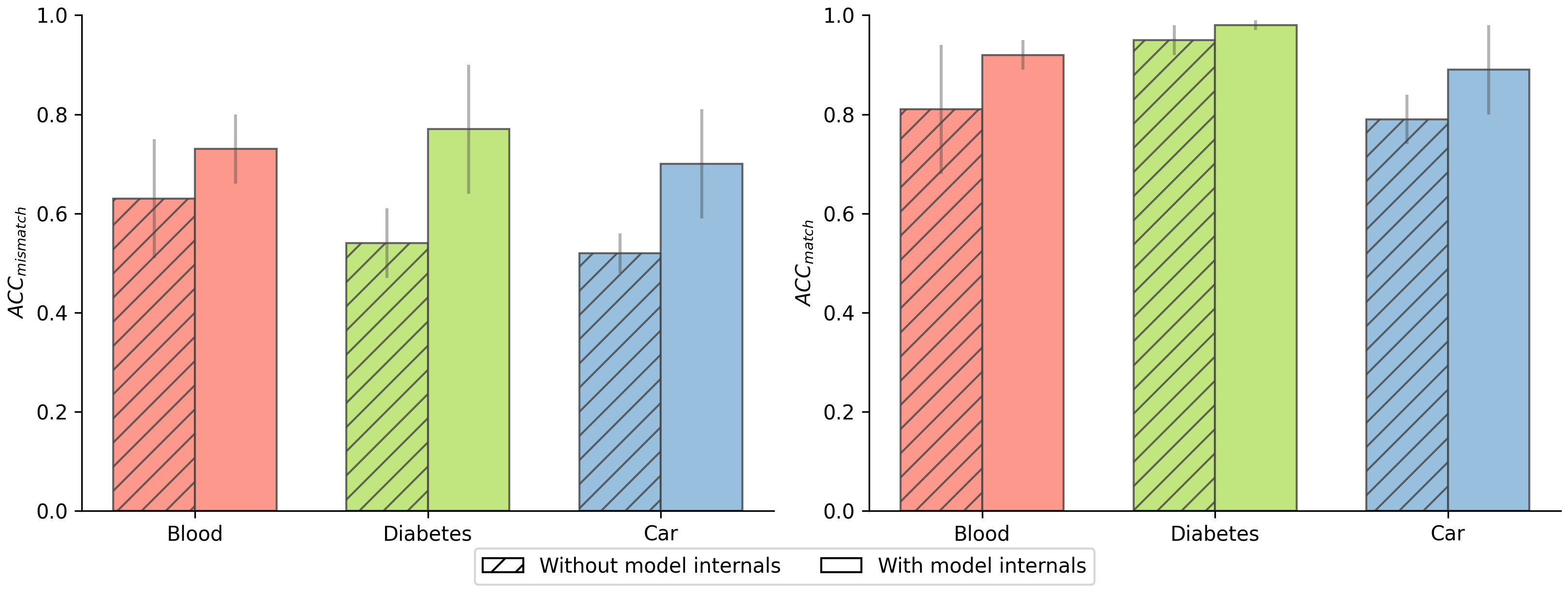}
    \caption{Comparison of GPT-4o's performance on DTs, with and without models' internals, for the \textcolor{spring_red}{Blood}, \textcolor{spring_green}{Diabetes}, and \textcolor{spring_blue}{Car} datasets. Including models' internals resulted in performance improvements across all cases.}
    \label{fig:DT_info_plot}
\end{figure}


For both LR and MLP, the inclusion of model internals led to performance either remaining within the margin of error or increasing modestly (typically 3-5\%) across all datasets. This suggests that while providing weights and structural information can help LLMs generate more accurate and faithful verbalizations, the relatively simple decision boundaries of LR and the black-box nature of MLPs may limit how much of this information the LLMs can meaningfully leverage. Notably, Llama and Gemini both showed the most consistent gains when MLP internals were included.

The most pronounced impact of incorporating model internals was observed for Decision Trees (\autoref{fig:DT_info_plot}). On the Blood dataset, GPT-4o's performance rose to a $\Accmismatch$ of $\accuracyWithError{0.73}{.07}$ and an $\Accmatch$ of $\accuracyWithError{0.92}{.03}$, marking a 15\% increase in $\Accoverall$. Claude's scores improved to a $\Accmismatch$ of $\accuracyWithError{0.70}{.02}$ and an $\Accmatch$ of $\accuracyWithError{0.91}{.01}$. Gemini also exhibited a notable jump, achieving a $\Accmismatch$ of $\accuracyWithError{0.75}{.12}$ and an $\Accmatch$ of $\accuracyWithError{.94}{.03}$. Similar improvements were observed across the remaining datasets, with all LLMs showing up to a 40\% increase in $\Accmismatch$ on the Diabetes dataset and around a 10\% gain in $\Accoverall$. The Car dataset also exhibited meaningful improvements.

These findings suggest that the rule-based nature of decision trees allowed LLMs to better capture and articulate the model's underlying decision logic. The explicit structure of decision paths in decision trees facilitated more accurate and interpretable verbalizations.

The impact of including model-specific information varied depending on the type of model. Only marginal improvements were observed in the scores for logistic regression and multilayer perceptron models. In contrast, decision trees witnessed the most substantial improvement, with performance gains across all datasets and all LLMs, with $\Accoverall$ even exceeding 0.9 in some cases. This underscores the effectiveness of including model-specific information in generating more accurate and faithful verbalizations. More broadly, our findings indicate that for certain model classes, access to internal structure can significantly enhance the quality of LLM-generated explanations.


\subsection{Ablation on Information on the Model's Type}
\label{subsec:ablation-model-type}
The model-type is the name of the type of model (e.g., Logistic Regression, Decision Tree, or MLP). We study the impact of excluding the model type when comparing models. We aim to evaluate if our framework can generate accurate verbalizations based purely on the observed behavior, rather than the names.

The Level 5 results in \autoref{add_results} show that removing model-type information from the prompt had little effect on the quality of verbalizations, with performance variations remaining within the margin of error. This implies that our framework relies mainly on the observed behavior (i.e., the representative sample) when verbalizing differences in decision boundaries.


It should be noted that all our ablation studies are conducted using stratification Level 2 (20-25\%) as the default configuration. Specific details about prompts can be found \autoref{prompts}.


\section{Discussion}
Our results show promising trends when verbalizing LR, DT, and MLP model differences. The non-linear nature of MLPs posed a greater challenge for our framework, as indicated by the decline in performance. Nevertheless, \model{} demonstrated reasonable effectiveness even in this more complex setting. This suggests that the framework can be extended to verbalizing the differences between DNNs, especially incorporating approaches that describe the models' internals (e.g., with mechanistic interpretability). Considering the complex nature of DNNs, the appropriate choice of a mechanistic interpretability approach will be crucial.

The plug-and-compare flexibility of \model{} allows potential upgrades. When newer, higher-capability LMs are developed, we can replace the LM in \model{} with the next-generation ones. The same flexibility applies to the prompting techniques and the expected tasks (for example, comparing across more than two models).

\model{} can be a foundational building block of a model resource manager. A good resource manager does not just observe; beyond verbalization, it should be able to automatically inspect the individual models, question the potential weaknesses, and potentially recommend improvement methods, including but not limited to model merging, model safeguarding, and model debiasing. Future work is needed toward this goal, which we believe deserves more attention from the field.
\section{Conclusion}


In conclusion, the \model{} framework establishes a foundational step toward the automated management and comparison of machine learning models. \model{} verbalizes the differences between two models by leveraging large language models (LLMs) to produce natural language descriptions, providing clear insights into the nuanced behaviors that performance metrics alone might not reveal. The framework is flexible and generalizable, accommodating various model types and datasets while also allowing the integration of additional model-specific information to enhance the accuracy and depth of the verbalizations.

Our experimental findings demonstrate that \model{} effectively captures behavioral differences between models across logistic regression, decision trees, and multilayer perceptrons. The experiments showed robust performance, achieving up to 80\% accuracy in verbalizing differences even when models had similar predictive performance but differed significantly in their behaviors. The inclusion of structural information, particularly in decision trees, markedly improved the informativeness of the generated verbalizations, with accuracy rising to around 90\% in these cases.

Model verbalizers are useful for future research in model management tools that can dynamically adapt to the evolving landscape of ML technologies. As we look to the future, integrating more sophisticated language models and expanding the framework's capabilities will be essential in advancing the field towards more transparent, accountable, and effective AI systems.

\section{Limitations}
While \model{} demonstrates promising results, several limitations remain that future research could address. Firstly, our evaluation was limited to logistic regression, decision trees, and MLP. Expanding the framework to include a broader range of models, such as deep neural networks with more layers and complex architectures, would further test and potentially enhance its applicability.

Additionally, we explored verbalizations using only a limited set of datasets. Future work should include a more diverse range of datasets, encompassing varying sizes, dimensionalities, and domains, to robustly assess the generalizability and effectiveness of the framework across different data contexts.

The current implementation of \model{} primarily utilizes basic structural information about the models. Including more comprehensive information—such as detailed performance metrics, accuracy differences, model complexity measures, and dataset-specific characteristics—could enhance the quality and depth of verbalizations.

Finally, we employed straightforward prompting techniques for the LLMs. Exploring advanced prompting strategies, such as iterative refinement prompts or in-context learning, could significantly improve the clarity, precision, and informativeness of the generated verbalizations. Addressing these limitations will be crucial for future iterations of \model{}, thereby moving closer to achieving comprehensive, automated model management and comparison.

\bibliographystyle{acl_natbib}
\bibliography{custom}


\newpage


\appendix

\section{Extended discussions}
\paragraph{Why not train an interpretable-by-design model?}
An alternative approach might involve training an interpretable model—e.g., a logistic regression—to approximate Model 2's predictions using the input features and Model 1's predictions. While this can reveal behavioral similarities and alignment between models, it differs fundamentally from our goal. Our framework focuses on generating natural language verbalizations that explain how two models differ, not on replicating one model's predictions through another. The evaluation framework employs a reconstruction-based method to assess how well the verbalization captures these differences, but this is strictly for evaluation purposes and is not part of the verbalization process itself. In contrast, interpretable-by-design models lack this linguistic component and address a different task altogether.

\paragraph{Explaining model disagreement induced by data variations.}
While our framework demonstrated its ability to generate meaningful verbalizations for differences in model predictions, previous experiments primarily focused on disagreements arising from systematic variations in model parameters. However, in real-world scenarios, prediction differences often emerged from diverse sources, including dataset-level perturbations and stochastic training variations. To evaluate the robustness of our approach under such conditions, we conducted an experiment where prediction disagreement was induced not through model changes, but through subtle modifications to the training data.

We used the Blood Donation dataset as our testbed and trained two models of the same type on distinct training sets. One model was trained on the original dataset, while the other was trained on a perturbed version, in which 25\% of the values in a single input feature (Recency) were randomly negated. This setup allowed us to assess the framework's performance in a scenario where model disagreements stemmed from dataset variations rather than architectural or parameter differences.

\autoref{tab:results-training-data-variations} (in Appendix \ref{subsec:results-training-data-variations}) reports the evaluation metrics across Logistic Regression (LR), Decision Tree (DT), and Multilayer Perceptron (MLP) models. Our framework continued to generate accurate and coherent verbalizations despite the absence of systematic model differences. For instance, GPT-4o and Claude maintained high $\Accmismatch$ scores (\accuracyWithError{0.82}{.01} and \accuracyWithError{0.88}{.01} on LR, respectively).  While performance varied across model types, with DTs showing slightly lower $\Accmismatch$ scores (e.g., Gemini achieved \accuracyWithError{0.71}{.08} on DTs), the framework still demonstrated robustness across different architectures. These results supported the generality of our approach in handling realistic, less structured sources of prediction disagreement.

\section{Additional Experimental Details}

\label{add_details:experiments}

\paragraph{DT generation:} For DTs, similar to the LR models, we first train a base model using \texttt{RandomizedSearchCV}. To generate a modified DT, we introduced variation at two levels. First, we randomly sample new hyperparameters from the defined space. This ensures that the modified tree has a structure different from the base model. Second, we add noise to the splitting thresholds of the trained tree. The noise was sampled from a normal distribution (\( \mathcal{N}(0, m) \)) and applied multiplicatively as \( \boldsymbol{\tau} \cdot (1 + \text{noise}) \), where \( \boldsymbol{\tau} \) denotes the original threshold values. We carefully adjust $m$ until the percentage of differing outputs between the base model and the modified model reaches the desired level. We also ensured that the difference in accuracy between any two models in a pair was $ \leq 5\%$, since in practice, models being compared would usually have similar performance.

\paragraph{MLP generation:} For MLPs, we first trained a base model using \texttt{RandomizedSearchCV}, keeping the architecture fixed and tuning hyperparameters such as the activation function, solver, and learning rate. To generate modified versions, we introduced noise into the model’s learned weights and biases. The noise was sampled from a normal distribution ($\mathcal{N}(0, m) $), where $m$ is the modification factor, and applied multiplicatively as $\mathbf{w} \cdot (1 + \text{noise})$, with $\mathbf{w}$  representing each parameter matrix or vector. We varied \( m \) to produce a range of model variations and selected pairs where the output disagreement fell within a target range and the difference in accuracy was \( \leq 5\% \). As with the other model families, we included both comparisons against the base model and comparisons between modified models to ensure a diverse set of model pairs.


\section{Prompts}

\label{prompts}


\subsection{Verbalization Prompts}


\begin{minipage}{\columnwidth}
\noindent\rule{\columnwidth}{0.4pt}

\begin{center}
\textbf{Verbalization Prompt} \\
(Blood Dataset)
\end{center}

\noindent\rule{\columnwidth}{0.4pt}
\vspace{0.5em}

\noindent
\small\ttfamily
\textbf{Context:} We have two  \{model\_type\} models trained on the same dataset for a binary classification task. The dataset contains details about random donors at a Blood Transfusion Service. The 4 features that it contains, in order, are: Recency (months), Frequency (times), Monetary (c.c. blood), and Time (months). The target feature (Blood Donated) is a binary variable representing whether the donor donated blood in March 2007 (1 stands for donating blood; 0 stands for not donating blood).\\

The sample dataset is provided below, which includes the 4 input features in the order mentioned above as well as the outputs/predictions generated by each of the two models.\\

\textbf{Dataset Sample:} \{verbalization\_data\}\\

\textbf{Task:} Your goal is to verbalize the differences between the decision boundaries of the two models based on the sample above.\\

\textbf{Instructions:} 
1. \textit{Analyze}: Examine the dataset sample to identify where the outputs of the two models diverge and where they align. \\ 
2. \textit{Quantify Divergences}: Determine the specific ranges of feature values (numerical intervals) where the decision boundaries of the two models diverge. Explain how the predictions differ within these ranges and where they remain consistent. Use numerical ranges for feature values to describe divergences; avoid vague terms like "high" or "low." \\
3. \textit{Identify Key Features}: Identify any features that significantly contribute to the observed differences in the models' outputs. \\
4. \textit{Verbalize Findings}: Provide a concise and precise verbal explanation of how the decision boundaries of the two models differ. Include numerical details to support your observations.\\

\textbf{Important Note:} Please do not include specific instances from the sample dataset in your response, as this could reveal sensitive information to a third person.
\end{minipage}


\noindent
\begin{minipage}[t]{0.48\textwidth}
\noindent\rule{\linewidth}{0.4pt}

\begin{center}
\textbf{Verbalization Prompt} \\
(Car Dataset)
\end{center}

\noindent\rule{\linewidth}{0.4pt}
\vspace{0.5em}

\noindent
\small\ttfamily
\textbf{Context:} We have two logistic regression models trained on the same dataset for a multiclass classification task. The dataset was derived from a simple hierarchical decision model developed for the evaluation of cars. The six features that it contains, in order, are: buying (buying price, encoding – \{"vhigh":3, "high": 2, "med": 1, "low": 0\}), maint (maintenance price, encoding – \{"vhigh": 3, "high": 2, "med": 1, "low": 0\}), doors (number of doors, encoding – \{"2": 2, "3": 3, "4": 4, "5more": 5\}), persons (capacity of persons, encoding – \{"2": 2, "4": 4, "more": 5\}), lug\_boot (size of luggage boot, encoding – \{"small": 0, "med": 1, "big": 2\}), and safety (safety level, encoding – \{"low": 0, "med": 1, "high": 2\}). The target variable (class) represents the overall evaluation of the car and has four possible values: 0 (Unacceptable), 1 (Acceptable), 2 (Good), 3 (Very Good).\\

The sample dataset is provided below, which includes the 6 input features in the order mentioned above as well as the outputs/predictions generated by each of the two models.\\

\textbf{Dataset Sample:} \{verbalization\_data\}\\

\textbf{Task:} Your goal is to verbalize the differences between the decision boundaries of the two models based on the sample above.\\

\textbf{Instructions:} 
1. \textit{Analyze}: Examine the dataset sample to identify where the outputs of the two models diverge and where they align.\\ 
2. \textit{Quantify Divergences}: Determine the specific ranges of feature values (numerical intervals) where the decision boundaries of the two models diverge. Explain how the predictions differ within these ranges and where they remain consistent. Use numerical ranges for feature values to describe divergences; avoid vague terms like "high" or "low."\\
3. \textit{Identify Key Features}: Identify any features that significantly contribute to the observed differences in the models' outputs.\\
4. \textit{Verbalize Findings}: Provide a concise and precise verbal explanation of how the decision boundaries of the two models differ. Include numerical details to support your observations.\\

\textbf{Important Note:} Please do not include specific instances from the sample dataset in your response, as this could reveal sensitive information to a third person.
\end{minipage}


\noindent
\begin{minipage}[t]{0.48\textwidth}
\noindent\rule{\linewidth}{0.4pt}

\begin{center}
\textbf{Verbalization Prompt} \\
(Diabetes Dataset)
\end{center}

\noindent\rule{\linewidth}{0.4pt}
\vspace{0.5em}

\noindent
\small\ttfamily
\textbf{Context:} We have two logistic regression models trained on the same dataset for a binary classification task. The objective of the dataset is to diagnostically predict whether or not a patient has diabetes, based on certain diagnostic measurements. The 8 features that it contains, in order, are: Pregnancies, Glucose, BloodPressure, SkinThickness, Insulin, BMI, DiabetesPedigreeFunction and Age. The target variable (Outcome) is a binary variable representing whether the patient has diabetes (1 stands for diabetic, 0 stands for non-diabetic) \\

The sample dataset is provided below, which includes the 8 input features in the order mentioned above as well as the outputs/predictions generated by each of the two models. \\

\textbf{Dataset Sample:} \{verbalization\_data\}\\

\textbf{Task:} Your goal is to verbalize the differences between the decision boundaries of the two models based on the sample above.\\

\textbf{Instructions:} 
1. \textit{Analyze}: Examine the dataset sample to identify where the outputs of the two models diverge and where they align.\\ 
2. \textit{Quantify Divergences}: Determine the specific ranges of feature values (numerical intervals) where the decision boundaries of the two models diverge. Explain how the predictions differ within these ranges and where they remain consistent. Use numerical ranges for feature values to describe divergences; avoid vague terms like "high" or "low."\\
3. \textit{Identify Key Features}: Identify any features that significantly contribute to the observed differences in the models' outputs.\\
4. \textit{Verbalize Findings}: Provide a concise and precise verbal explanation of how the decision boundaries of the two models differ. Include numerical details to support your observations.\\

\textbf{Important Note:} Please do not include specific instances from the sample dataset in your response, as this could reveal sensitive information to a third person.
\end{minipage}

\noindent
\begin{minipage}[t]{0.48\textwidth}
\noindent\rule{\linewidth}{0.4pt}

\begin{center}
\textbf{Verbalization Prompt} \\
(Blood Dataset With Model Information)
\end{center}

\noindent\rule{\linewidth}{0.4pt}
\vspace{0.5em}

\noindent
\small\ttfamily
\textbf{Context:} We have two  \{model\_type\} models trained on the same dataset for a binary classification task. The dataset contains details about random donors at a Blood Transfusion Service. The 4 features that it contains, in order, are: Recency (months), Frequency (times), Monetary (c.c. blood), and Time (months). The target feature (Blood Donated) is a binary variable representing whether the donor donated blood in March 2007 (1 stands for donating blood; 0 stands for not donating blood).\\

The sample dataset is provided below, which includes the 4 input features in the order mentioned above as well as the outputs/predictions generated by each of the two models.\\

\textbf{Dataset Sample:} \{verbalization\_data\}\\

\textbf{Model Information:} The weights for each logistic regression model are provided below. These weights correspond to the input features in the same order as described above: \{model\_info\} \\

\textbf{Task:} Your goal is to verbalize the differences between the decision boundaries of the two models based on the sample and model information above. \\

\textbf{Instructions:} 
1. \textit{Analyze}: Examine the dataset sample to identify where the outputs of the two models diverge and where they align. Review the given model information to understand how and why the models might produce different outputs. \\
2. \textit{Quantify Divergences}: Determine the specific ranges of feature values (numerical intervals) where the decision boundaries of the two models diverge. Explain how the predictions differ within these ranges and where they remain consistent. Use numerical ranges for feature values to describe divergences; avoid vague terms like "high" or "low." \\
3. \textit{Identify Key Features}: Identify any features that significantly contribute to the observed differences in the models' outputs. \\
4. \textit{Verbalize Findings}: Provide a concise and precise verbal explanation of how the decision boundaries of the two models differ. Include numerical details to support your observations.\\

\textbf{Important Note:} Please do not include specific instances from the sample dataset in your response, as this could reveal sensitive information to a third person.
\end{minipage}


\begin{minipage}[t]{0.48\textwidth}
\noindent\rule{\linewidth}{0.4pt}

\begin{center}
\textbf{Evaluation Prompt} \\
(Blood Dataset)
\end{center}

\noindent\rule{\linewidth}{0.4pt}
\vspace{0.5em}

\noindent
\small\ttfamily
\textbf{Context}: We have two \{model\_type\} models trained on the same dataset for a binary classification task. The dataset contains details about random donors at a Blood Transfusion Service. The 4 features that it contains, in order, are: Recency (months), Frequency (times), Monetary (c.c. blood), and Time (months). The target feature (Blood Donated) is a binary variable representing whether the donor donated blood in March 2007 (1 stands for donating blood; 0 stands for not donating blood).\\

Below is a sample of the dataset, which includes the 4 input features in the order mentioned above and the outputs/predictions generated by \{model\_unpruned\}. The accompanying verbalization provides a verbal explanation of how the decision boundaries of the two models differ. \\

\textbf{Dataset Sample:} \texttt{\{evaluation\_data\}}\\

\textbf{Verbalization:} \texttt{\{verbalization\}}\\

\textbf{Task:} Based on the verbalization, predict the output of \{model\_pruned\} for each instance in the sample above. \\

\textbf{Instructions:} Think about the question carefully. Go through the verbalization thoroughly. Analyze the input features in the sample. After explaining your reasoning, provide the answer in a JSON format within markdown at the end. The JSON should contain the input features and the output of \{model\_pruned\}. The format of the JSON should be the same as the dataset sample. Do not provide any further details after the JSON.\\

\end{minipage}


\noindent
\begin{minipage}[t]{0.48\textwidth}
\noindent\rule{\linewidth}{0.4pt}

\begin{center}
\textbf{Evaluation Prompt} \\
(Car Dataset)
\end{center}

\noindent\rule{\linewidth}{0.4pt}
\vspace{0.5em}

\noindent
\small\ttfamily
\textbf{Context}: We have two logistic regression models trained on the same dataset for a multiclass classification task. The dataset was derived from a simple hierarchical decision model developed for the evaluation of cars. The six features that it contains, in order, are: buying (buying price, encoding – \{"vhigh":3, "high": 2, "med": 1, "low": 0\}), maint (maintenance price, encoding – \{"vhigh": 3, "high": 2, "med": 1, "low": 0\}), doors (number of doors, encoding – \{"2": 2, "3": 3, "4": 4, "5more": 5\}), persons (capacity of persons, encoding – \{"2": 2, "4": 4, "more": 5\}), lug\_boot (size of luggage boot, encoding – \{"small": 0, "med": 1, "big": 2\}), and safety (safety level, encoding – \{"low": 0, "med": 1, "high": 2\}). The target variable (class) represents the overall evaluation of the car and has four possible values: 0 (Unacceptable), 1 (Acceptable), 2 (Good), 3 (Very Good).\\

Below is a sample of the dataset, which includes the 6 input features in the order mentioned above and the outputs/predictions generated by \{model\_pruned\}. The accompanying verbalization provides a verbal explanation of how the decision boundaries of the two models differ. \\

\textbf{Dataset Sample:} \texttt{\{evaluation\_data\}}\\

\textbf{Verbalization:} \texttt{\{verbalization\}}\\

\textbf{Task:} Based on the verbalization, predict the output of \{model\_pruned\} for each instance in the sample above. \\

\textbf{Instructions:} Think about the question carefully. Go through the verbalization thoroughly. Analyze the input features in the sample. After explaining your reasoning, provide the answer in a JSON format within markdown at the end. The JSON should contain the input features and the output of \{model\_pruned\}. The format of the JSON should be the same as the dataset sample. Do not provide any further details after the JSON.\\

\end{minipage}


\noindent
\begin{minipage}[t]{0.48\textwidth}
\noindent\rule{\linewidth}{0.4pt}

\begin{center}
\textbf{Evaluation Prompt} \\
(Diabetes Dataset)
\end{center}

\noindent\rule{\linewidth}{0.4pt}
\vspace{0.5em}

\noindent
\small\ttfamily
\textbf{Context}: We have two \{model\_type\} models trained on the same dataset for a binary classification task. The objective of the dataset is to diagnostically predict whether or not a patient has diabetes, based on certain diagnostic measurements. The 8 features that it contains, in order, are: Pregnancies, Glucose, BloodPressure, SkinThickness, Insulin, BMI, DiabetesPedigreeFunction and Age. The target variable (Outcome) is a binary variable representing whether the patient has diabetes (1 stands for diabetic, 0 stands for non-diabetic) \\

Below is a sample of the dataset, which includes the 8 input features in the order mentioned above and the outputs/predictions generated by \{model\_unpruned\}. The accompanying verbalization provides a verbal explanation of how the decision boundaries of the two models differ. \\

\textbf{Dataset Sample:} \texttt{\{evaluation\_data\}}\\

\textbf{Verbalization:} \texttt{\{verbalization\}}\\

\textbf{Task:} Based on the verbalization, predict the output of \{model\_pruned\} for each instance in the sample above. \\

\textbf{Instructions:} Think about the question carefully. Go through the verbalization thoroughly. Analyze the input features in the sample. After explaining your reasoning, provide the answer in a JSON format within markdown at the end. The JSON should contain the input features and the output of \{model\_pruned\}. The format of the JSON should be the same as the dataset sample. Do not provide any further details after the JSON.\\

\end{minipage}

\onecolumn

\section{Additional Experiment Results}

\label{add_results}

\subsection{Excerpts from model verbalization}

\begin{table}[h]
\centering
\renewcommand{\arraystretch}{1.5} 
\resizebox{\linewidth}{!}{
\begin{tabular}{l p{16cm} }
\toprule
\textbf{Model} & \textbf{Example Verbalization Excerpts} \\ \midrule
Claude & \ldots The divergence is most pronounced when Recency is in the -90 to -65 range, Frequency and Monetary are very low (around -75 to -60), and Time is very low (-135 to -80). In these scenarios, Model 1 predicts a positive outcome, while Model 2 predicts a negative outcome \ldots \\ \midrule
GPT & \ldots The decision boundaries of the two logistic regression models diverge primarily in the negative ranges of the features Recency, Frequency, Monetary, and Time. Specifically, when the values of these features are within the ranges of approximately -90 to -60 for Recency, -77 to -61 for Frequency and Monetary, and -135 to -126 for Time, Model 2 is more likely to predict a positive outcome (1) compared to Model 1, which tends to predict a negative outcome (0) \ldots \\ \midrule
Gemini & \ldots Model 2 appears to predict '1' (donating blood) more often than Model 1 for certain ranges of feature values. Specifically, Model 2 predicts '1' when: \\
& - \textbf{Frequency} is positive and \textbf{Time} is between approximately -25 and 40 months, and \textbf{Recency} is less than approximately -70 months \ldots \\ 
\bottomrule

\end{tabular}}
\caption{Verbalization Excerpts for a pair of Level 2 ($20-25\%$) LR models trained on the Blood Dataset.}
\label{tab:verbalization-lr}
\end{table}

\begin{table}[h]
\centering
\captionsetup{justification=centering}
\renewcommand{\arraystretch}{1.5} 
\resizebox{\linewidth}{!}{
\begin{tabular}{l p{16cm} }
\toprule
\textbf{Model} & \textbf{Example Verbalization Excerpts} \\ \midrule
Claude & \ldots Model 2 tends to be more lenient in its classifications compared to Model 1, often assigning higher class values. \\

& The most significant divergence occurs when the buying price (first feature) is 0 or 1, maintenance price (second feature) is 0 or 1, and safety (sixth feature) is 2. In these cases, Model 2 often assigns a class of 2, while Model 1 assigns a class of 0 or 1 \ldots \\ \midrule
GPT & \ldots When the buying price is low (0), Model 2 tends to predict higher classes (more favorable evaluations) compared to Model 1, especially when other conditions such as safety, capacity, and maintenance are moderate to high. \\

& Similar to buying price, lower maintenance costs (0) often lead to higher predictions in Model 2 compared to Model 1, under favorable conditions in other features \ldots \\ \midrule

Gemini & \ldots **Model 2 demonstrates greater leniency towards cars with higher maintenance costs.** Specifically, when the maintenance price is "high" (encoded as 2) or "very high" (encoded as 3), Model 2 tends to assign a higher evaluation class compared to Model 1, even when other features like buying price are less favorable \ldots \\ \bottomrule
\end{tabular}}
\caption{Verbalization Excerpts for a pair of Level 2 ($20-25\%$) DT models trained on the Car Dataset.}
\label{tab:verbalization-dt}
\end{table}

\newpage

\subsection{Numerical Results}


\begin{table}[ht]
\centering
\captionsetup{skip=10pt}
\footnotesize
\renewcommand{\arraystretch}{1.0}
\begin{tabular}{llccccc}
\toprule


\textbf{LLM} & \textbf{Metric} & \textbf{Level 1} & \textbf{Level 2} & \textbf{Level 3} & \textbf{Level 4} & \textbf{Level 5} \\
\midrule
\multicolumn{7}{l}{\textbf{Blood Dataset}} \\
\midrule
\multirow{3}{*}{GPT-4o} 
  & $\Accoverall$  & 0.78 ± .10 & 0.74 ± .08 & 0.73 ± .09 & 0.77 ± .07 & 0.76 ± .06 \\
  & $\Accmismatch$ & 0.76 ± .20 & 0.70 ± .06 & 0.67 ± .05 & 0.72 ± .08 & 0.69 ± .12 \\
  & $\Accmatch$    & 0.78 ± .08 & 0.75 ± .09 & 0.75 ± .11 & 0.77 ± .10 & 0.77 ± .08 \\
\midrule
\multirow{3}{*}{Claude 3.5 Sonnet}
  & $\Accoverall$  & 0.79 ± .03 & 0.79 ± .04 & 0.77 ± .06 & 0.79 ± .04 & 0.79 ± .04 \\
  & $\Accmismatch$ & 0.81 ± .17 & 0.67 ± .10 & 0.70 ± .04 & 0.68 ± .12 & 0.69 ± .09 \\
  & $\Accmatch$    & 0.78 ± .05 & 0.81 ± .06 & 0.79 ± .08 & 0.81 ± .07 & 0.80 ± .06 \\
\midrule
\multirow{3}{*}{Gemini 2.0 Flash}
  & $\Accoverall$  & 0.84 ± .03 & 0.80 ± .04 & 0.81 ± .04 & 0.81 ± .07 & 0.82 ± .05 \\
  & $\Accmismatch$ & 0.78 ± .22 & 0.64 ± .13 & 0.70 ± .04 & 0.66 ± .15 & 0.63 ± .10 \\
  & $\Accmatch$    & 0.85 ± .02 & 0.83 ± .03 & 0.85 ± .05 & 0.84 ± .05 & 0.85 ± .04 \\
\midrule
\multirow{3}{*}{Llama 3.3}
  & $\Accoverall$  & 0.82 ± .07 & 0.73 ± .15 & 0.74 ± .09 & 0.75 ± .10 & 0.74 ± .09 \\
  & $\Accmismatch$ & 0.66 ± .20 & 0.72 ± .11 & 0.56 ± .07 & 0.72 ± .13 & 0.73 ± .12 \\
  & $\Accmatch$    & 0.85 ± .05 & 0.74 ± .16 & 0.80 ± .09 & 0.76 ± .09 & 0.75 ± .08 \\


\toprule
\multicolumn{7}{l}{\textbf{Diabetes Dataset}} \\
\midrule
\multirow{3}{*}{GPT-4o} 
  & $\Accoverall$  & 0.82 ± .02 & 0.79 ± .08 & 0.76 ± .03 & 0.81 ± .05 & 0.78 ± .04 \\
  & $\Accmismatch$ & 0.56 ± .06 & 0.58 ± .15 & 0.56 ± .12 & 0.58 ± .06 & 0.55 ± .05 \\
  & $\Accmatch$    & 0.91 ± .04 & 0.87 ± .05 & 0.83 ± .08 & 0.90 ± .06 & 0.88 ± .06 \\
\midrule
\multirow{3}{*}{Claude 3.5 Sonnet}
  & $\Accoverall$  & 0.82 ± .06 & 0.81 ± .04 & 0.75 ± .05 & 0.80 ± .06 & 0.79 ± .04 \\
  & $\Accmismatch$ & 0.58 ± .07 & 0.57 ± .04 & 0.57 ± .08 & 0.56 ± .08 & 0.54 ± .05 \\
  & $\Accmatch$    & 0.89 ± .03 & 0.89 ± .04 & 0.83 ± .08 & 0.89 ± .05 & 0.89 ± .04 \\
\midrule
\multirow{3}{*}{Gemini 2.0 Flash}
  & $\Accoverall$  & 0.75 ± .05 & 0.79 ± .07 & 0.67 ± .06 & 0.80 ± .08 & 0.78 ± .06 \\
  & $\Accmismatch$ & 0.56 ± .14 & 0.61 ± .10 & 0.59 ± .16 & 0.62 ± .11 & 0.61 ± .09 \\
  & $\Accmatch$    & 0.79 ± .04 & 0.86 ± .09 & 0.71 ± .08 & 0.87 ± .10 & 0.85 ± .07 \\
\midrule
\multirow{3}{*}{Llama 3.3}
  & $\Accoverall$  & 0.69 ± .06 & 0.71 ± .07 & 0.69 ± .06 & 0.74 ± .04 & 0.71 ± .05 \\
  & $\Accmismatch$ & 0.56 ± .05 & 0.55 ± .05 & 0.54 ± .05 & 0.57 ± .06 & 0.54 ± .04 \\
  & $\Accmatch$    & 0.72 ± .07 & 0.78 ± .09 & 0.75 ± .06 & 0.81 ± .05 & 0.79 ± .05 \\


\toprule
\multicolumn{7}{l}{\textbf{Car Dataset}} \\
\midrule
\multirow{3}{*}{GPT-4o} 
  & $\Accoverall$  & 0.70 ± .05 & 0.75 ± .06 & 0.69 ± .04 & 0.79 ± .07 & 0.73 ± .05 \\
  & $\Accmismatch$ & 0.54 ± .08 & 0.54 ± .01 & 0.52 ± .14 & 0.57 ± .05 & 0.51 ± .07 \\
  & $\Accmatch$    & 0.74 ± .07 & 0.82 ± .08 & 0.75 ± .02 & 0.86 ± .10 & 0.81 ± .06 \\
\midrule
\multirow{3}{*}{Claude 3.5 Sonnet}
  & $\Accoverall$  & 0.69 ± .03 & 0.75 ± .05 & 0.71 ± .09 & 0.75 ± .06 & 0.75 ± .04 \\
  & $\Accmismatch$ & 0.56 ± .17 & 0.49 ± .07 & 0.56 ± .13 & 0.51 ± .10 & 0.46 ± .08 \\
  & $\Accmatch$    & 0.73 ± .03 & 0.84 ± .06 & 0.77 ± .09 & 0.83 ± .06 & 0.87 ± .05 \\
\midrule
\multirow{3}{*}{Gemini 2.0 Flash}
  & $\Accoverall$  & 0.66 ± .04 & 0.71 ± .05 & 0.66 ± .03 & 0.70 ± .08 & 0.74 ± .06 \\
  & $\Accmismatch$ & 0.56 ± .12 & 0.56 ± .11 & 0.53 ± .12 & 0.56 ± .10 & 0.58 ± .07 \\
  & $\Accmatch$    & 0.69 ± .08 & 0.77 ± .08 & 0.71 ± .03 & 0.76 ± .07 & 0.80 ± .06 \\
\midrule
\multirow{3}{*}{Llama 3.3}
  & $\Accoverall$  & 0.62 ± .05 & 0.69 ± .06 & 0.60 ± .08 & 0.69 ± .08 & 0.67 ± .07 \\
  & $\Accmismatch$ & 0.56 ± .09 & 0.53 ± .05 & 0.50 ± .09 & 0.53 ± .05 & 0.53 ± .06 \\
  & $\Accmatch$    & 0.64 ± .08 & 0.75 ± .10 & 0.64 ± .11 & 0.75 ± .09 & 0.72 ± .08 \\

\bottomrule
\end{tabular}
\caption{Evaluation metrics for LR models across different datasets. Each row includes the performance metrics for an LLM, measured across Level 1 ($15-20\%$), Level 2 ($20-25\%$), Level 3 ($25-30\%$), Level 4 ($20-25\%$ With Models' Internals), and Level 5 ($20-25\%$ Without Model Type).}
\label{tab:lr_extend}
\end{table}


\begin{table}[ht]
\centering
\captionsetup{skip=10pt}
\footnotesize
\renewcommand{\arraystretch}{1.0}
\begin{tabular}{llccccc}
\toprule


\textbf{LLM} & \textbf{Metric} & \textbf{Level 1} & \textbf{Level 2} & \textbf{Level 3} & \textbf{Level 4} & \textbf{Level 5} \\
\midrule
\multicolumn{7}{l}{\textbf{Blood Dataset}} \\
\midrule
\multirow{3}{*}{GPT-4o} 
  & $\Accoverall$  & 0.83 ± .03 & 0.77 ± .09 & 0.75 ± .04 & 0.88 ± .03 & 0.79 ± .08 \\
  & $\Accmismatch$ & 0.67 ± .09 & 0.63 ± .12 & 0.69 ± .10 & 0.73 ± .07 & 0.65 ± .12 \\
  & $\Accmatch$    & 0.87 ± .02 & 0.81 ± .13 & 0.76 ± .03 & 0.92 ± .03 & 0.83 ± .06 \\
\midrule                                                         
\multirow{3}{*}{Claude 3.5 Sonnet}                               
  & $\Accoverall$  & 0.83 ± .07 & 0.77 ± .04 & 0.81 ± .07 & 0.86 ± .02 & 0.76 ± .05 \\
  & $\Accmismatch$ & 0.70 ± .10 & 0.65 ± .08 & 0.66 ± .14 & 0.70 ± .02 & 0.67 ± .06 \\
  & $\Accmatch$    & 0.86 ± .09 & 0.81 ± .05 & 0.85 ± .04 & 0.91 ± .01 & 0.79 ± .05 \\
\midrule                                                         
\multirow{3}{*}{Gemini 2.0 Flash}                                
  & $\Accoverall$  & 0.84 ± .03 & 0.83 ± .02 & 0.80 ± .06 & 0.90 ± .05 & 0.84 ± .04 \\
  & $\Accmismatch$ & 0.67 ± .10 & 0.70 ± .06 & 0.71 ± .13 & 0.75 ± .12 & 0.70 ± .05 \\
  & $\Accmatch$    & 0.89 ± .02 & 0.86 ± .03 & 0.83 ± .06 & 0.94 ± .03 & 0.87 ± .04 \\
\midrule                                                         
\multirow{3}{*}{Llama 3.3}                                       
  & $\Accoverall$  & 0.77 ± .08 & 0.80 ± .04 & 0.79 ± .06 & 0.83 ± .02 & 0.79 ± .04 \\
  & $\Accmismatch$ & 0.67 ± .03 & 0.55 ± .08 & 0.64 ± .07 & 0.75 ± .03 & 0.54 ± .07 \\
  & $\Accmatch$    & 0.79 ± .09 & 0.86 ± .04 & 0.83 ± .06 & 0.85 ± .02 & 0.86 ± .05 \\


\toprule
\multicolumn{7}{l}{\textbf{Diabetes Dataset}} \\
\midrule
\multirow{3}{*}{GPT-4o} 
  & $\Accoverall$  & 0.85 ± .02 & 0.82 ± .02 & 0.84 ± .02 & 0.91 ± 0.07 & 0.81 ± .03 \\
  & $\Accmismatch$ & 0.53 ± .11 & 0.54 ± .07 & 0.67 ± .06 & 0.77 ± 0.13 & 0.53 ± .06 \\
  & $\Accmatch$    & 0.94 ± .02 & 0.95 ± .03 & 0.95 ± .02 & 0.98 ± 0.01 & 0.93 ± .03 \\
\midrule                                                         
\multirow{3}{*}{Claude 3.5 Sonnet}                               
  & $\Accoverall$  & 0.84 ± .02 & 0.81 ± .03 & 0.78 ± .02 & 0.90 ± 0.07 & 0.80 ± .03 \\
  & $\Accmismatch$ & 0.57 ± .05 & 0.50 ± .08 & 0.55 ± .03 & 0.74 ± 0.15 & 0.50 ± .04 \\
  & $\Accmatch$    & 0.93 ± .04 & 0.96 ± .04 & 0.92 ± .03 & 0.98 ± 0.01 & 0.95 ± .03 \\
\midrule                                                         
\multirow{3}{*}{Gemini 2.0 Flash}                                
  & $\Accoverall$  & 0.78 ± .02 & 0.78 ± .05 & 0.80 ± .01 & 0.91 ± 0.05 & 0.77 ± .04 \\
  & $\Accmismatch$ & 0.53 ± .14 & 0.56 ± .08 & 0.55 ± .07 & 0.82 ± 0.11 & 0.57 ± .06 \\
  & $\Accmatch$    & 0.85 ± .02 & 0.89 ± .08 & 0.95 ± .04 & 0.94 ± 0.04 & 0.86 ± .05 \\
\midrule                                                         
\multirow{3}{*}{Llama 3.3}                                       
  & $\Accoverall$  & 0.75 ± .05 & 0.75 ± .05 & 0.67 ± .09 & 0.89 ± 0.07 & 0.76 ± .04 \\
  & $\Accmismatch$ & 0.51 ± .11 & 0.51 ± .03 & 0.54 ± .05 & 0.75 ± 0.13 & 0.52 ± .04 \\
  & $\Accmatch$    & 0.81 ± .04 & 0.85 ± .06 & 0.75 ± .13 & 0.96 ± 0.02 & 0.87 ± .05 \\


\toprule
\multicolumn{7}{l}{\textbf{Car Dataset}} \\
\midrule
\multirow{3}{*}{GPT-4o} 
  & $\Accoverall$  & 0.71 ± .04 & 0.73 ± .02 & 0.67 ± .08 & 0.85 ± .08 & 0.72 ± .03 \\
  & $\Accmismatch$ & 0.43 ± .06 & 0.52 ± .04 & 0.48 ± .10 & 0.70 ± .11 & 0.52 ± .05 \\
  & $\Accmatch$    & 0.77 ± .04 & 0.79 ± .05 & 0.75 ± .06 & 0.89 ± .09 & 0.78 ± .04 \\
\midrule                                                         
\multirow{3}{*}{Claude 3.5 Sonnet}                               
  & $\Accoverall$  & 0.64 ± .07 & 0.74 ± .04 & 0.62 ± .10 & 0.80 ± .09 & 0.73 ± .03 \\
  & $\Accmismatch$ & 0.49 ± .13 & 0.57 ± .07 & 0.46 ± .04 & 0.67 ± .06 & 0.57 ± .06 \\
  & $\Accmatch$    & 0.68 ± .10 & 0.78 ± .04 & 0.67 ± .12 & 0.84 ± .09 & 0.78 ± .05 \\
\midrule                                                         
\multirow{3}{*}{Gemini 2.0 Flash}                                
  & $\Accoverall$  & 0.72 ± .06 & 0.68 ± .07 & 0.62 ± .11 & 0.85 ± .09 & 0.68 ± .06 \\
  & $\Accmismatch$ & 0.54 ± .05 & 0.58 ± .13 & 0.54 ± .11 & 0.66 ± .11 & 0.59 ± .09 \\
  & $\Accmatch$    & 0.76 ± .08 & 0.71 ± .09 & 0.65 ± .11 & 0.89 ± .07 & 0.71 ± .08 \\
\midrule                                                         
\multirow{3}{*}{Llama 3.3}                                       
  & $\Accoverall$  & 0.60 ± .17 & 0.67 ± .07 & 0.57 ± .17 & 0.80 ± .08 & 0.66 ± .06 \\
  & $\Accmismatch$ & 0.62 ± .03 & 0.58 ± .08 & 0.51 ± .12 & 0.70 ± .13 & 0.57 ± .05 \\
  & $\Accmatch$    & 0.60 ± .21 & 0.69 ± .09 & 0.58 ± .21 & 0.82 ± .08 & 0.69 ± .06 \\

\bottomrule
\end{tabular}
\caption{Evaluation metrics for DT models across different datasets. Each row includes the performance metrics for an LLM, measured across Level 1 ($15-20\%$), Level 2 ($20-25\%$), Level 3 ($25-30\%$), Level 4 ($20-25\%$ With Models' Internals), and Level 5 ($20-25\%$ Without Model Type).}
\label{tab:dt_extend}
\end{table}


\begin{table}[ht]
\centering
\captionsetup{skip=10pt}
\footnotesize
\renewcommand{\arraystretch}{1.0}
\begin{tabular}{llccccc}
\toprule


\textbf{LLM} & \textbf{Metric} & \textbf{Level 1} & \textbf{Level 2} & \textbf{Level 3} & \textbf{Level 4} & \textbf{Level 5} \\
\midrule
\multicolumn{7}{l}{\textbf{Blood Dataset}} \\
\midrule
\multirow{3}{*}{GPT-4o} 
  & $\Accoverall$  & 0.82 ± .08 & 0.80 ± .03 & 0.72 ± .07 & 0.81 ± .04 & 0.79 ± .04 \\
  & $\Accmismatch$ & 0.67 ± .15 & 0.66 ± .10 & 0.56 ± .07 & 0.65 ± .09 & 0.69 ± .08 \\
  & $\Accmatch$    & 0.85 ± .06 & 0.83 ± .04 & 0.77 ± .06 & 0.85 ± .04 & 0.82 ± .05 \\
\midrule                                                         
\multirow{3}{*}{Claude 3.5 Sonnet}                               
  & $\Accoverall$  & 0.80 ± .04 & 0.81 ± .04 & 0.70 ± .07 & 0.81 ± .06 & 0.84 ± .04 \\
  & $\Accmismatch$ & 0.72 ± .15 & 0.62 ± .09 & 0.62 ± .09 & 0.65 ± .08 & 0.64 ± .06 \\
  & $\Accmatch$    & 0.82 ± .07 & 0.85 ± .04 & 0.73 ± .07 & 0.84 ± .07 & 0.88 ± .05 \\
\midrule                                                         
\multirow{3}{*}{Gemini 2.0 Flash}                                
  & $\Accoverall$  & 0.77 ± .05 & 0.82 ± .04 & 0.80 ± .02 & 0.86 ± .03 & 0.84 ± .03 \\
  & $\Accmismatch$ & 0.73 ± .14 & 0.66 ± .09 & 0.63 ± .04 & 0.68 ± .08 & 0.67 ± .06 \\
  & $\Accmatch$    & 0.78 ± .09 & 0.86 ± .05 & 0.86 ± .02 & 0.90 ± .02 & 0.88 ± .04 \\
\midrule                                                         
\multirow{3}{*}{Llama 3.3}                                       
  & $\Accoverall$  & 0.81 ± .02 & 0.82 ± .01 & 0.74 ± .06 & 0.87 ± .02 & 0.80 ± .04 \\
  & $\Accmismatch$ & 0.66 ± .08 & 0.62 ± .09 & 0.56 ± .06 & 0.61 ± .02 & 0.60 ± .11 \\
  & $\Accmatch$    & 0.84 ± .02 & 0.87 ± .01 & 0.80 ± .05 & 0.93 ± .01 & 0.85 ± .05 \\


\toprule
\multicolumn{7}{l}{\textbf{Diabetes Dataset}} \\
\midrule
\multirow{3}{*}{GPT-4o} 
  & $\Accoverall$  & 0.82 ± .02 & 0.81 ± .03 & 0.78 ± .03 & 0.81 ± .04 & 0.79 ± .03 \\
  & $\Accmismatch$ & 0.63 ± .08 & 0.55 ± .04 & 0.60 ± .09 & 0.63 ± .02 & 0.55 ± .04 \\
  & $\Accmatch$    & 0.88 ± .02 & 0.90 ± .03 & 0.85 ± .03 & 0.88 ± .05 & 0.88 ± .04 \\
\midrule                                                         
\multirow{3}{*}{Claude 3.5 Sonnet}                               
  & $\Accoverall$  & 0.81 ± .02 & 0.82 ± .01 & 0.78 ± .05 & 0.82 ± .01 & 0.80 ± .03 \\
  & $\Accmismatch$ & 0.60 ± .05 & 0.57 ± .09 & 0.51 ± .10 & 0.57 ± .04 & 0.58 ± .05 \\
  & $\Accmatch$    & 0.87 ± .02 & 0.90 ± .03 & 0.87 ± .01 & 0.90 ± .02 & 0.88 ± .03 \\
\midrule                                                         
\multirow{3}{*}{Gemini 2.0 Flash}                                
  & $\Accoverall$  & 0.79 ± .03 & 0.79 ± .02 & 0.74 ± .05 & 0.80 ± .02 & 0.78 ± .03 \\
  & $\Accmismatch$ & 0.58 ± .04 & 0.58 ± .07 & 0.59 ± .17 & 0.62 ± .07 & 0.58 ± .06 \\
  & $\Accmatch$    & 0.86 ± .02 & 0.86 ± .01 & 0.81 ± .05 & 0.86 ± .04 & 0.84 ± .03 \\
\midrule                                                         
\multirow{3}{*}{Llama 3.3}                                       
  & $\Accoverall$  & 0.77 ± .03 & 0.73 ± .05 & 0.69 ± .08 & 0.81 ± .04 & 0.74 ± .04 \\
  & $\Accmismatch$ & 0.58 ± .05 & 0.53 ± .07 & 0.56 ± .07 & 0.60 ± .06 & 0.55 ± .05 \\
  & $\Accmatch$    & 0.82 ± .05 & 0.80 ± .10 & 0.72 ± .09 & 0.87 ± .06 & 0.81 ± .05 \\


\toprule
\multicolumn{7}{l}{\textbf{Car Dataset}} \\
\midrule
\multirow{3}{*}{GPT-4o} 
  & $\Accoverall$  & 0.74 ± .02 & 0.68 ± .05 & 0.66 ± .02 & 0.63 ± .10 & 0.68 ± .04 \\
  & $\Accmismatch$ & 0.59 ± .07 & 0.51 ± .10 & 0.52 ± .09 & 0.63 ± .05 & 0.52 ± .05 \\
  & $\Accmatch$    & 0.76 ± .03 & 0.74 ± .05 & 0.72 ± .02 & 0.64 ± .13 & 0.73 ± .06 \\
\midrule                                                         
\multirow{3}{*}{Claude 3.5 Sonnet}                               
  & $\Accoverall$  & 0.71 ± .03 & 0.60 ± .08 & 0.65 ± .08 & 0.62 ± .13 & 0.59 ± .05 \\
  & $\Accmismatch$ & 0.58 ± .04 & 0.47 ± .15 & 0.58 ± .07 & 0.61 ± .02 & 0.46 ± .06 \\
  & $\Accmatch$    & 0.75 ± .03 & 0.64 ± .08 & 0.69 ± .10 & 0.65 ± .17 & 0.62 ± .07 \\
\midrule                                                         
\multirow{3}{*}{Gemini 2.0 Flash}                                
  & $\Accoverall$  & 0.70 ± .08 & 0.67 ± .07 & 0.68 ± .05 & 0.71 ± .09 & 0.69 ± .04 \\
  & $\Accmismatch$ & 0.59 ± .18 & 0.58 ± .13 & 0.64 ± .09 & 0.70 ± .05 & 0.60 ± .07 \\
  & $\Accmatch$    & 0.72 ± .08 & 0.70 ± .06 & 0.69 ± .07 & 0.71 ± .13 & 0.72 ± .06 \\
\midrule                                                         
\multirow{3}{*}{Llama 3.3}                                       
  & $\Accoverall$  & 0.63 ± .11 & 0.62 ± .07 & 0.60 ± .04 & 0.67 ± .11 & 0.62 ± .05 \\
  & $\Accmismatch$ & 0.55 ± .08 & 0.52 ± .14 & 0.53 ± .07 & 0.71 ± .09 & 0.52 ± .06 \\
  & $\Accmatch$    & 0.64 ± .11 & 0.64 ± .05 & 0.63 ± .06 & 0.64 ± .13 & 0.65 ± .06 \\

\bottomrule
\end{tabular}
\caption{Evaluation metrics for MLP models across different datasets. Each row includes the performance metrics for an LLM, measured across Level 1 ($15-20\%$), Level 2 ($20-25\%$), Level 3 ($25-30\%$), Level 4 ($20-25\%$ With Models' Internals), and Level 5 ($20-25\%$ Without Model Type).}
\label{tab:mlp_extend}
\end{table}

\clearpage
\subsection{Results on differences induced by training data variations}
\label{subsec:results-training-data-variations}
\vspace{30pt}

\begin{table}[ht]
\centering
\captionsetup{skip=10pt}
\footnotesize
\renewcommand{\arraystretch}{1.0}
\begin{tabular}{llccccc}
\toprule


\textbf{LLM} & \textbf{Metric} & \textbf{LR} & \textbf{DT} & \textbf{MLP} \\
\midrule
\multicolumn{5}{l}{\textbf{Blood Dataset}} \\
\midrule
\multirow{3}{*}{GPT-4o} 
  & $\Accoverall$  & 0.70 ± .07 & 0.84 ± .04 & 0.79 ± .06  \\
  & $\Accmismatch$ & 0.82 ± .01 & 0.65 ± .15 & 0.64 ± .08  \\
  & $\Accmatch$    & 0.75 ± .04 & 0.85 ± .03 & 0.81 ± .06  \\
\midrule                                                  
\multirow{3}{*}{Claude 3.5 Sonnet}                        
  & $\Accoverall$  & 0.80 ± .05 & 0.88 ± .03 & 0.81 ± .05  \\
  & $\Accmismatch$ & 0.88 ± .01 & 0.64 ± .13 & 0.64 ± .16  \\
  & $\Accmatch$    & 0.75 ± .09 & 0.90 ± .05 & 0.82 ± .05  \\
\midrule                                                  
\multirow{3}{*}{Gemini 2.0 Flash}                         
  & $\Accoverall$  & 0.81 ± .03 & 0.83 ± .07 & 0.87 ± .04  \\
  & $\Accmismatch$ & 0.96 ± .02 & 0.71 ± .08 & 0.60 ± .14  \\
  & $\Accmatch$    & 0.70 ± .08 & 0.84 ± .08 & 0.89 ± .06  \\
\midrule                                                  
\multirow{3}{*}{Llama 3.3}                                
  & $\Accoverall$  & 0.76 ± .04 & 0.69 ± .09 & 0.79 ± .05  \\
  & $\Accmismatch$ & 0.67 ± .04 & 0.64 ± .11 & 0.67 ± .13  \\
  & $\Accmatch$    & 0.83 ± .03 & 0.70 ± .10 & 0.81 ± .05  \\

\bottomrule
\end{tabular}
\caption{Evaluation metrics for LLMs on explanations of prediction differences induced by training data variations. We report performance metrics for each underlying model -- Logistic Regression (LR), Decision Tree (DT), and Multilayer Perceptron (MLP).}
\
\label{tab:results-training-data-variations}
\end{table}

\end{document}